\documentclass[10pt, a4paper]{article}

\usepackage[final]{lrec-coling2024} 

\usepackage{color}
\usepackage{multirow} 
\usepackage{makecell} 
\usepackage{algorithm}
\usepackage{algorithmic}
\usepackage{amsmath}
\usepackage{float}
\usepackage{capt-of}

\title{Enhancing Effectiveness and Robustness in a Low-Resource Regime via Decision-Boundary-aware Data Augmentation}

\name{Kyohoon Jin\textsuperscript{$^\ast$1}\thanks{$^\ast$Equal contribution}, 
      Junho Lee\textsuperscript{$^\ast$2}, 
      Juhwan Choi\textsuperscript{$^\ast$2}, 
      Sangmin Song\textsuperscript{$^\ast$2}, 
      Youngbin Kim\textsuperscript{$^\dagger$1,2}\thanks{$^\dagger$Corresponding author}} 

\address{\\ \textsuperscript{1}Graduate School of Advanced Imaging Sciences, Multimedia and Film, Chung-Ang University \\
         \textsuperscript{2}Department of Artificial Intelligence, Chung-Ang University \\
         \{fhzh123, jhjo32, gold5230, s2022120859, ybkim85\}@cau.ac.kr\\}

\abstract{
Efforts to leverage deep learning models in low-resource regimes have led to numerous augmentation studies. However, the direct application of methods such as mixup and cutout to text data, is limited due to their discrete characteristics. While methods using pretrained language models have exhibited efficiency, they require additional considerations for robustness. Inspired by recent studies on decision boundaries, this paper proposes a decision-boundary-aware data augmentation strategy to enhance robustness using pretrained language models. The proposed technique first focuses on shifting the latent features closer to the decision boundary, followed by reconstruction to generate an ambiguous version with a soft label. Additionally, mid-K sampling is suggested to enhance the diversity of the generated sentences. This paper demonstrates the performance of the proposed augmentation strategy compared to other methods through extensive experiments. Furthermore, the ablation study reveals the effect of soft labels and mid-K sampling and the extensibility of the method with curriculum data augmentation.
 \\ \newline  }

\begin{document}

\maketitleabstract

\section{Introduction}
As the latest pretrained language models have demonstrated excellent performance, numerous studies have been conducted on training larger models with more data. However, due to the numerous parameters that need to be learned, these pretrained language models require considerable data for downstream tasks. Data augmentation is widely used to address this problem, preventing overfitting by increasing the quantity of training data. Consequently, various data augmentation methods have been studied across various fields, including computer vision, audio, and text \cite{rizos2019augment, lee2021crossaug, oh2021influence, lee2022efficient}. 

Various studies have proposed data augmentation methods that transform the data while preserving their attributes as much as possible, such as rotation and Cutout~\cite{devries2017improved}. For textual data, fundamental textual operations, such as replacement, insertion, deletion, and shuffling, have been widely accepted in various augmentation frameworks \cite{wei-zou-2019-eda}. This straightforward data augmentation strategy enhances model robustness by focusing the optimization process on strengthening its ability to handle noise~\cite{neelakantan2015adding,piedboeuf-langlais-2022-effective}. The introduction of noise increases robustness, enabling the model to maintain its performance against intentional text corruption or modifications~\cite{karpukhin-etal-2019-training}. However, these techniques face the challenge of not guaranteeing the preservation of attributes and readability between the original and augmented sentences~\cite{hsu-etal-2021-semantics}. Methods using pretrained language models trained on diverse data to learn language representations have been proposed for data augmentation to address the preservation of attributes and readability problems. Compared to the noise addition, these models offer better preservation of readability and attributes \cite{wiechmann-kerz-2022-measuring}. Nevertheless, they have limited ability to provide various sentences as they generate sentences based on the existing data distribution \cite{pmlr-v80-ott18a, vanmassenhove-etal-2019-lost}. Furthermore, as these sentences are generated using language models pretrained on massive data, they tend to generate typical expressions. Moreover, the data is generated without consideration of decision boundaries, leading to a lack of robustness \cite{dong2021should}.

On the other hand, One of the popular data augmentation techniques is mixup, which involves combining two or more pieces of information from different images to create a new image. Mixup configures the vicinal risk minimization learning method using soft labels instead of one-hot encoding labels in the learning process. This approach helps prevent overconfidence, enhances robustness against adversarial attacks, and preserves the content of each attribute~\cite{zhang2018Mixup}. It offers several benefits, such as enhancing model robustness against adversarial attacks by training the decision boundary using soft labels, as found in previous studies \cite{9754227}. However, directly applying mixup-based approaches to the text domain is limited. Unlike images, where attributes can be interpreted as continuous signals, sentences consist of a discrete set of words. Consequently, the modification of words in equal ratios does not guarantee an equal influence on the sentence label \cite{kim2021linda, chen-etal-2022-doublemix}.

From a geometric perspective, the robustness of a deep learning model is expected to be influenced by its decision boundary \cite{goodfellow2015explaining}. Decision boundaries extend across the entire feature space used for training and are not limited to the provided data points. Therefore, investigating the decision boundaries is a crucial aspect of understanding the decision-making behavior of deep neural network classifiers.

Accordingly, we propose a data augmentation technique that leverages the advantages of each method. Similar to mixup, the proposed approach involves shifting the latent features toward the decision boundary, leveraging soft labels while effectively preserving existing attributes. We define ambiguous data as values close to the decision boundary. Furthermore, we employ a pretrained language model to ensure readability and attribute consistency between the original and augmented sentences. Additionally, we introduce variability to sentences through mid-K sampling and enhance robustness through decision-boundary-aware gradient modification. Experiments demonstrate that the proposed method improves the performance of the model by constructing a more robust decision boundary by shifting it using augmented data. Additionally, the experimental result showcased the superiority of our method compared to previous baselines in terms of performance enhancement, statistical durability, and robustness against adversarial attacks.

Our contributions are summarized as follows.

\begin{itemize}

\item We propose a novel and intuitive data augmentation technique considering the decision boundary in latent space. In addition, we exploit a pretrained language model to preserve the readability and attribute consistency in the generated sentences, enhancing the effectiveness of the data augmentation.

\item We introduce mid-K sampling to generate diverse augmented data. This method generates diverse data by selecting the top $k$ words while considering the middle $K$ words, generating data while preserving essential information.

\item We demonstrate the effectiveness and robustness of the proposed method through comparative experiments including soft labels, curriculum learning, and various decoding strategies.

\end{itemize}

\section{Related Work}
\subsection{Text Augmentation}

Text data augmentation is a training strategy designed to enhance the robustness of a model by generating new sentences using various techniques. One common approach is the manipulation of words according to predefined rules. Easy data augmentation (EDA) is a well-known method that employs rule-based techniques, including synonym replacement, random insertion, swap, and deletion, to introduce diverse types of noise \cite{wei-zou-2019-eda}. However, these techniques might alter the original meaning of the sentences by randomly deleting words or changing their order without considering the context. In contrast to word-level rule-based augmentation, other methods have been proposed to consider the context of sentences using deep learning models. 

Early research in this area employed language models to replace words with their alternatives, considering the context of the sentence \cite{kobayashi-2018-contextual, wu2019conditional, zhou-etal-2022-flipda}. These model-based augmentation techniques preserve the semantics of the original data by modifying only a portion of the original sentence. However, they may have limited diversity compared to the original data. Further developing the existing research, a study proposed using a pretrained large language model (LLM) \cite{yoo-etal-2021-gpt3mix-leveraging}. However, this method requires existing LLM knowledge and may result in bias.

On the other hand, several approaches have been proposed to introduce mixup augmentation into the natural language processing field \cite{zhang2018Mixup}. Early studies suggested performing mixup interpolation at the word embedding level or the level of encoded sentence representations \cite{guo2020nonlinear}. An end-to-end scheme to perform mixup and train the model was also introduced \cite{sun-etal-2020-Mixup, chen-etal-2022-doublemix}. These mixup-based approaches complement the sparsity of the data distributions through interpolation. However, they have limited interpretability because the augmented results cannot be directly observed.

\begin{figure*}[ht]
    \centering
    \includegraphics[width=0.9\textwidth]{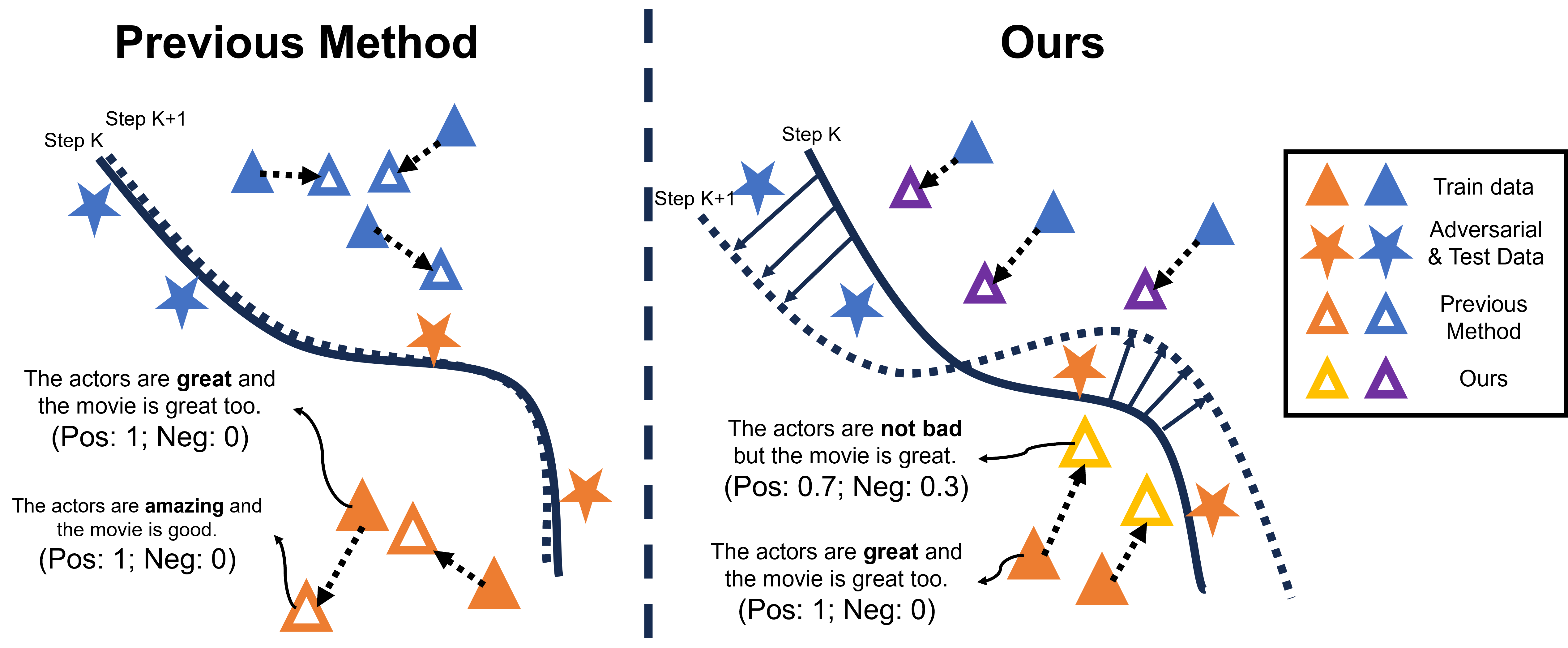}
    \caption{The figure illustrates the concept of the decision-boundary-aware gradient modification. In the previous method, augmentation was performed without the consideration of decision boundaries. However, in the proposed method, augmentation is performed in decision-boundary-aware manner.}
    \label{fig1:overview}
\end{figure*}

\subsection{Analysis of Decision-Boundary Neighbored Data}

In training and evaluating deep learning models, various studies have suggested the importance of counterfactual data, which have minimal differences in the input space but different label values \cite{teney2020learning, gardner-etal-2020-evaluating, Kaushik2020Learning}. Specifically, the concept of a contrast set has been introduced to explain this phenomenon from the viewpoint of the decision boundary \cite{gardner-etal-2020-evaluating}. Ambiguous data that are close to the decision boundary and challenging to distinguish are essential in forming a robust decision boundary \cite{Ding2020MMA, zhu-etal-2022-improving} and in training \cite{margatina-etal-2021-active, zhang-etal-2022-allsh}. Data points close to the decision boundary are also significant from the perspective of adversarial attacks because vulnerability to adversarial assaults increases with the distance from the decision boundary \cite{zhang2021geometryaware}. Considering this geometric characteristic, strategies to improve the generalizability of a model by adjusting the perturbation have been studied \cite{zhang2022tactics, holtz2022learning, ijcai2022p0095, yang2022one, zhang2023delving}.

Previous studies have demonstrated the potential to enhance model robustness through data augmentation \cite{rebuffi2021data, shorten2021text, gowal2021improving}. Based on these findings, the concept of decision-boundary-aware data augmentation \cite{zhu-etal-2022-improving, 9754227} has been emerged. Decision-Boundary-aware augmentation method aims to establish a more robust decision boundary by leveraging data augmentation techniques that can shift the current decision boundary in a targeted direction \cite{park-etal-2022-consistency, peng2023generating}. A previous study suggested a method to identify adversarial samples that lie close to the decision boundary without crossing it, improving adversarial robustness without sacrificing performance \cite{zhu-etal-2022-improving}.



Furthermore, studies have explored the utilization of mixup to generate data points close to the decision boundary. However, existing mixup augmentation methods are claimed to be ineffective in improving adversarial robustness when interpolation samples are randomly selected \cite{9754227}. To address this problem, researchers have proposed a method that explicitly selects the current attackable data as mixup interpolation samples to generate attackable data points. However, in the field of natural language processing, the generation of meaningful sentences is difficult because of the discrete representation of text and variations in length \cite{yoon-etal-2021-ssmix}. Thus, previous studies that applied mixup augmentation in the natural language processing domain have primarily focused on the feature level \cite{guo2020nonlinear, sun-etal-2020-Mixup}. To overcome this limitation, SSMix~\cite{yoon-etal-2021-ssmix} suggested generating sentences with soft labels by modifying words that influence the label. However, these studies involve word-level modifications and differ from the direct generation of sentences that combine two sets of label information. Inspired by these studies, we construct sentences with soft labels by reconstructing them from ambiguous representations after shifting the encoded representations to the decision boundary.

\section{Methodology}
\subsection{Task Definition}
In this paper, the decision boundary refers to the region where the probability of each class is equal \cite{karimi2022decision}. We defined ambiguous data as values close to the decision boundary. A task is defined as obtaining $\hat{x}$, corresponding to generating ambiguous data from the given source data $x$, and each data $x$ is paired with an attribute vector $y$. For example, in sentiment analysis, sentiments such as 'positive' and 'negative' become attributes $y$. The data augmentation process consists of four steps. First, we create a well-trained attribute classifier $C_{\pi}$ with source data $x$ and source attribute $y$. Second, the source data $x$ is encoded based on the encoder from the first step to obtain $z$, which is the latent representation of $x$. To examine the similarity of $z$ to the decision boundary in the latent space, we pass $z$ to $C_{\pi}$ to obtain the classification $\tilde{y}$. Third, based on the gradient of $\tilde{y}$ for the decision boundary, we apply $n$ times the iterative gradient modification to the value of $z$ to obtain an ambiguous representation $z'^{n}$. The Transformer-based decoder reconstructs $x$ from this $z'^{n}$, resulting in $\hat{x}$, the augmented data of $x$. Finally, $\hat{x}$ is input into $C_{\pi}$ for scoring, and the result is assigned as a soft label to create an augmented data pair $D' = \{\hat{x}, \hat{y} \}$ in the result. Figure~\ref{fig1:overview} describes the overall proposed task.

\subsection{Model Training}
The proposed model consists of three subcomponents: the encoder, decoder, and attribute classifier. The entire procedure to augment data is divided into four steps. First, the encoder $E_{\theta}$ encodes a given sentence into its latent representation $z$. Then, the attribute classifier is trained by $z$, resulting in a well-trained attribute classifier. In addition, the encoder learns how to precisely distinguish each attribute in the latent space. The training object is defined as follows:

\begin{gather}
\begin{aligned}
&\mathcal{L}_{cls}(C_{\pi}(E_{\theta}(x), y; \theta, \pi)) = \\
&\varepsilon_{cls}\sum_{i}^{|C|}{u_i}log(q_i) - (1-\varepsilon_{cls})\sum_{i}^{|C|}\bar{q_i}log(q_i)
\end{aligned}
\end{gather}

Second, the frozen $E_{\theta}$ generates $z$, which is used by the Transformer-decoder $D_{\gamma}$ to conduct reconstruction from the provided $z$. Through this process, $D_{\gamma}$ is trained to reconstruct a sentence from $z$. The training object is defined as follows:

\begin{gather}
\begin{aligned}
\mathcal{L}_{recon}(D_{\gamma}(E_{\theta}(x), x; \gamma)) = \\
-\sum_{k=1}^{|N|}\sum^{|x_k|} \bigg( (1-\varepsilon_{recon})\sum_i^{|V|}\bar{p_i}log(p_i) \\
+ \varepsilon_{recon}\sum_i^{|V|}\bar{u_i}log(p_i) \bigg)
\end{aligned}
\end{gather}


where $|N|$ represents the size of the training data, $|x_k|$ represents the length of $x_k$, and $|V|$ and $|C|$ denote the size of the vocabulary and number of classes, respectively. In addition, $\bar{p}_i$ and $\bar{q}_i$ represent the probability distributions predicted by the decoder and classifier, respectively. Further, $p_i$ and $q_i$ represent the true distribution of reconstruction and classification, respectively. Moreover, $\varepsilon_{recon}$ and $\varepsilon_{cls}$ are label smoothing parameters for each loss term, and $u_i$ represents a uniform noise distribution for label smoothing, defined as $1/|V|$ and $1/|C|$, respectively.

During training, we first trained $E_{\theta}$ and $C_{\pi}$ using $\mathcal{L}_{cls}$ and then trained $D_{\gamma}$ using $\mathcal{L}_{recon}$ while fixing $E_{\theta}$. Thus, $C_{\pi}$ and $D_{\gamma}$ are trained separately, independent of each other. We found that separate training of $C_{\pi}$ and $D_{\gamma}$ yields superior results compared to joint training. Following this approach to model training, the decision-boundary-aware gradient is modified to provide enhanced data $\hat{x}$.

\subsection{Decision Boundary-aware Gradient Modification} 

During the inference time (i.e., augmenting a given sentence), $D_{\gamma}$ takes $z'^{(n)}$, the ambiguous representation of $z'=E_{\theta}(x)$ as input. To acquire the transformed representation, we passed $z$ to $C_{\pi}$ and computed the gradient. The direction of modification is determined by back-propagating the gradient of the attribute classification loss, inspired by previous work \cite{goodfellow2014explaining}. During the iterative modification process, instead of using $z$, we adapted the previous step $z'^{(n-1)}$. In addition, $\lambda$ is a hyperparameter for modification: 

\begin{gather}
\begin{aligned}
z'^{(n)}=z'^{(n-1)}-\lambda \bigtriangledown_{z'^(n-1)} \mathcal{L}_{cls}(C_{\pi}(z'^{(n-1)}), \bar{y}) \\ where\;z'^{(0)}:= z,\;n >= 1 
\end{aligned}
\end{gather}

where $\bar{y}$ indicates the decision boundary of the model. The decision boundary is defined as every class with equal likelihood (e.g., $\{0.5, 0.5\}$ for a binary classification task).

\begin{figure}[t]
    \centering
    \includegraphics[width=0.9\columnwidth]{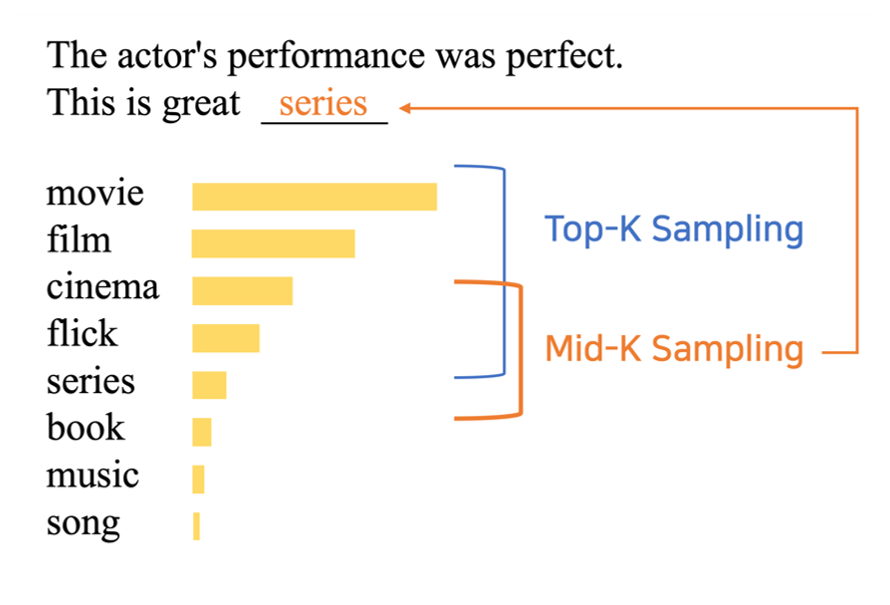}
    \caption{The figure illustrates the concept of the mid-K Sampling. Mid-K Sampling is a method to increase the diversity of generated sentences by sampling the middle K sentences instead of selecting the K sentences with the highest probability values (Top-K Sampling).}
    \label{fig2:midk}
\end{figure}

\subsection{Mid-K Sampling}

Although reconstruction was performed from a modified representation, the difference between generated and original sentences may not be noticeable, depending on the decoding strategy such as beam search~\cite{freitag-al-onaizan-2017-beam}. Existing decoding strategies, such as top-K \cite{fan-etal-2018-hierarchical} and nucleus sampling \cite{Holtzman2020The}, aim to generate sentences close to the correct sentence. Although this characteristic is useful for general tasks, such as machine translation, which requires an optimal solution, it may be improper for data augmentation, which requires diverse generated sentences \cite{feng-etal-2021-survey}. We propose a novel technique called mid-K sampling to achieve the goal of the augmentation method, which generates sentences that differ from the original while preserving the core semantics.

Figure~\ref{fig2:midk} illustrates the concept of mid-K sampling. Similar to top‑K sampling, mid-K sampling selects the $k$ most probable words at the current time step to exclude inappropriate words that are too different from the original sentence. Next, we further selected the word $k'$ with the highest probability among the $k$ words. To consider the significance of the selected $k'$ words, we determine d whether the cumulative sum of the probabilities of these words exceeds the predefined threshold $p$.

If the cumulative sum is below the threshold, it indicates that the word distribution is relatively flat at this time step. Therefore, the words generated in this step are less crucial in determining the meaning of the sentence, allowing a variety of word selections. By explicitly excluding the most likely $k'$ words from the distribution, it is possible to generate an ambiguous sentence that differs from the original and prevents the uniformity of the generated sentences.

Conversely, if the cumulative sum exceeds the threshold, the word distribution has a high skewness. In this case, the $k'$ words can be expected to be crucial in maintaining the meaning of the sentence. Therefore, sampling is performed among all $k$ words, including $k'$ words, which is the same as top‑K sampling. By allowing critical words to be generated, the core semantics of the original sentence can be preserved. 

\begin{algorithm}
\caption{The Procedure of Mid-K Sampling}
\begin{algorithmic}[1]
\REQUIRE Word probability distribution $p$, Vocabulary $V$, $k$, $k'$, threshold $t$
\ENSURE Next word token
\STATE Get top $k$ token from distribution $p$
\STATE Normalize word probability with respect to $k$ tokens and get $p_k$
\STATE Get the cumulative sum of the probability of top $k'$ tokens from $p_k$
\IF{cumsum $\geq$ $t$}
    \STATE Get $p_{k'}$ by excluding $k'$ tokens from $p_k$ and normalize
    \STATE Sample next token from $p_{k'}$
\ELSE
    \STATE Sample next token from $p_k$
\ENDIF
\end{algorithmic}
\end{algorithm}

\section{Experiments}

\subsection{Experimental Setup}

We conducted the experiments on various text classification datasets \cite{socher-etal-2013-recursive, warstadt-etal-2019-neural, pang-etal-2002-thumbs, li-roth-2002-learning, maas-etal-2011-learning} and baselines \cite{wei-zou-2019-eda, wu2019conditional, sennrich-etal-2016-improving, anaby2020not, yoo-etal-2021-gpt3mix-leveraging}. Additionally, to simulate more challenging scenarios in learning the model, we evaluated the proposed method under low-resource conditions using only a subset of the training set for each dataset. We repeated the same experiment five times with different random seeds and reported the mean value and standard deviation of the results. We adopted \textit{t5-large} from Transformers library \cite{wolf-etal-2020-transformers} as the encoder and decoder of our model. The evaluation of downstream tasks was performed using the \textit{bert-base-cased} model and the \textit{microsoft/deberta-v3-base} model from Transformers.

\begin{table*}[t]
\centering
\resizebox{\textwidth}{!}{%
\begin{tabular}{cc|ccccccc}
\Xhline{3\arrayrulewidth}
\multicolumn{2}{c|}{\textbf{BERT}}    & Baseline    & EDA                               & BT                                & C-BERT                            & LAMBADA                           & GPT3-MIX                           & Ours                \\ \Xhline{2\arrayrulewidth}
                            & 1\%     & 82.0 (2.8)  & {\color[HTML]{858585} 79.6 (1.9)} & {\color[HTML]{858585} 80.7 (3.1)} & 83.1 (2.7)                        & 84.7 (4.4)                        & \textit{87.7 (0.6)}                & \textbf{88.9 (0.6)} \\
\multirow{-2}{*}{SST2}      & 10\%    & 89.4 (6.1)  & {\color[HTML]{858585} 88.2 (2.4)} & {\color[HTML]{858585} 86.9 (4.5)} & {\color[HTML]{858585} 88.9 (4.1)} & \textit{90.1 (3.6)}               & {\color[HTML]{858585} 86.2 (0.5)}  & \textbf{90.5 (1.6)} \\ \hline
                            & 1\%     & 61.7 (3.0)  & 63.2 (4.7)                        & {\color[HTML]{858585} 56.6 (6.4)} & 62.9 (7.3)                        & \textbf{68.9 (3.9)}               & \textit{68.5 (0.3)}                & 67.5 (2.0)          \\
\multirow{-2}{*}{CoLA}      & 10\%    & 64.7 (6.3)  & 68.3 (4.1)                        & 69.2 (4.1)                        & {\color[HTML]{858585} 63.6 (7.1)} & 67.6 (4.0)                        & \textit{69.4 (2.2)}                & \textbf{69.7 (0.6)} \\ \hline
                            & 1\%     & 73.4 (11.8) & {\color[HTML]{858585} 72.7 (4.3)} & 73.5 (4.2)                        & {\color[HTML]{858585} 72.3 (4.3)} & 85.5 (1.5)                        & \textbf{90.6 (1.1)}                & \textit{86.7 (1.1)}    \\
\multirow{-2}{*}{SUBJ}      & 10\%    & 77.7 (10.2) & 80.2 (5.7)                        & {\color[HTML]{858585} 76.5 (7.7)} & 80.2 (14.4)                       & 87.4 (6.7)                        & \textit{91.1 (1.2)}                & \textbf{91.7 (3.9)} \\ \hline
                            & 1\%     & 26.5 (4.1)  & {\color[HTML]{858585} 25.8 (3.6)} & {\color[HTML]{858585} 25.8 (2.4)} & {\color[HTML]{858585} 26.4 (2.9)} & 29.4 (2.6)                        & \textbf{33.3 (0.1)}                & \textit{30.4 (1.6)}    \\
\multirow{-2}{*}{SST5}      & 10\%    & 44.4 (4.5)  & 46.6 (1.5)                        & 46.5 (2.9)                        & 47.6 (5.1)                        & \textit{48.5 (4.5)}               & {\color[HTML]{858585} 43.0 (1.8)}  & \textbf{49.5 (1.7)} \\ \hline
                            & 1\%     & 67.0 (7.5)  & {\color[HTML]{858585} 65.9 (7.1)} & \textbf{69.3 (6.3)}               & {\color[HTML]{858585} 66.6 (5.9)} & {\color[HTML]{858585} 63.2 (4.9)} & {\color[HTML]{858585} 60.5 (6.1)}  & \textit{68.2 (4.1)}    \\
\multirow{-2}{*}{TREC6}     & 10\%    & 83.4 (3.9)  & 84.1 (5.4)                        & 86.1 (7.7)                        & 86.3 (7.5)                        & {\color[HTML]{858585} 71.3 (6.9)} & \textit{86.6 (2.7)}                & \textbf{88.3 (1.6)} \\ \hline
                            & 1\%     & 70.0 (4.1)  & {\color[HTML]{858585} 65.6 (4.8)} & {\color[HTML]{858585} 66.9 (3.3)} & 78.2 (0.1)                        & 80.7 (6.7)                        & \textbf{86.2 (1.8)}                & \textit{84.6 (5.7)}    \\
\multirow{-2}{*}{IMDB}      & 10\%    & 81.6 (5.3)  & {\color[HTML]{858585} 67.3 (3.0)} & {\color[HTML]{858585} 77.5 (5.9)} & {\color[HTML]{858585} 75.2 (5.3)} & \textit{82.1 (1.5)}               & -                                  & \textbf{87.7 (1.9)} \\ \Xhline{3\arrayrulewidth}
\multicolumn{2}{c|}{\textbf{DeBERTa}} & Baseline    & EDA                               & BT                                & C-BERT                            & LAMBADA                           & GPT3-MIX                           & Ours                \\ \Xhline{2\arrayrulewidth}
                            & 1\%     & 81.4 (5.6)  & 85.2 (4.9)                        & 85.5 (6.6)                        & 84.5 (6.6)                        & 87.5 (5.3)                        & \textit{88.2 (1.3)}                & \textbf{88.6 (1.1)} \\
\multirow{-2}{*}{SST2}      & 10\%    & 86.5 (6.7)  & 89.2 (3.2)                        & {\color[HTML]{858585} 85.5 (6.6)} & 91.6 (1.4)                        & 92.5 (1.8)                        & \textit{93.2 (0.8)}                & \textbf{93.4 (0.9)} \\ \hline
                            & 1\%     & 66.5 (3.7)  & 67.4 (3.7)                        & 69.2 (5.2)                        & 69.2 (5.2)                        & 69.0 (0.5)                        & \textit{75.4 (4.4)}                & \textbf{79.9 (3.1)} \\
\multirow{-2}{*}{CoLA}      & 10\%    & 79.0 (2.7)  & {\color[HTML]{858585} 65.6 (4.5)} & {\color[HTML]{858585} 66.4 (4.7)} & {\color[HTML]{858585} 73.9 (5.2)} & \textbf{80.5 (6.5)}               & 79.0 (1.3)                         & \textit{79.3 (2.5)}    \\ \hline
                            & 1\%     & 77.7 (7.8)  & 78.0 (3.2)                        & 78.7 (7.9)                        & {\color[HTML]{858585} 75.8 (4.3)} & 80.2 (2.4)                        & \textbf{88.6 (2.1)}                & \textit{83.7 (1.8)}    \\
\multirow{-2}{*}{SUBJ}      & 10\%    & 83.1 (6.5)  & 86.3 (7.1)                        & 85.8 (7.1)                        & 83.7 (8.9)                        & \textit{88.2 (5.2)}               & {\color[HTML]{858585} 82.9 (14.7)} & \textbf{89.6 (2.2)} \\ \hline
                            & 1\%     & 26.4 (2.4)  & 26.7 (6.7)                        & {\color[HTML]{858585} 24.7 (4.5)} & 26.7 (2.8)                        & 26.9 (1.5)                        & \textbf{37.5 (4.0)}                & \textit{29.2 (0.3)}    \\
\multirow{-2}{*}{SST5}      & 10\%    & 25.8 (4.0)  & {\color[HTML]{858585} 25.6 (3.2)} & 26.2 (2.9)                        & \textit{28.6 (4.8)}               & \textbf{29.9 (5.4)}               & 27.4 (11.0)                        & \textbf{29.9 (0.7)} \\ \hline
                            & 1\%     & 61.0 (3.8)  & 64.4 (4.4)                        & 65.5 (3.4)                        & 66.0 (2.9)                        & \textit{67.0 (2.7)}               & 63.6 (11.8)                        & \textbf{68.1 (4.1)} \\
\multirow{-2}{*}{TREC6}     & 10\%    & 90.1 (6.1)  & {\color[HTML]{858585} 86.7 (7.0)} & 90.4 (6.3)                        & 93.6 (2.6)                        & 93.4 (5.7)                        & \textit{94.2 (0.9)}                & \textbf{95.6 (1.2)} \\ \hline
                            & 1\%     & 77.8 (2.1)  & {\color[HTML]{858585} 77.0 (6.3)} & {\color[HTML]{858585} 74.8 (5.5)} & 87.4 (7.3)                        & {\color[HTML]{858585} 70.3 (3.5)} & \textit{87.7 (1.8)}                & \textbf{89.9 (2.2)} \\
\multirow{-2}{*}{IMDB}      & 10\%    & 86.0 (3.5)  & {\color[HTML]{858585} 84.8 (4.9)} & {\color[HTML]{858585} 85.1 (4.3)} & \textit{88.0 (3.0)}               & {\color[HTML]{858585} 83.9 (3.2)} & -                                  & \textbf{91.9 (2.1)} \\ \Xhline{3\arrayrulewidth}
\end{tabular}%
}
\caption{Performance (\%) of baseline and proposed method on the BERT~\cite{devlin-etal-2019-bert} and DeBERTa~\cite{he2021deberta} models. The statistics are presented in the mean (standard deviation) format, and 1\% and 10\% indicate what percentage of the original was used. The best results are boldfaced, and the second-best results are italicized. Lower scores than the baseline are in gray. For the Internet Movie Database (IMDB) dataset, we could not experiment with the IMDB-10\% environment due to the limitation of the official source code.}
\label{tab:main_table}
\end{table*}

\subsection{Effectiveness}

Table~\ref{tab:main_table} presents the experimental results. This study conducts experiments with BERT, which is commonly used, and DeBERTa to evaluate the effectiveness of the proposed method on larger models. The results reveal that the proposed method demonstrates a high average performance gain compared to other augmentation methods in the majority of datasets.

Both EDA \cite{wei-zou-2019-eda} and back-translation \cite{sennrich-etal-2016-improving}, which are widely used as existing data augmentation techniques, performed similarly or even led to performance degradation compared to the baseline in BERT and DeBERTa. This phenomenon may arise from two factors: the potential semantic damage in EDA and the lack of diversity in the augmented data generated by back-translation.

The use of pretrained language models and the incorporation of soft labels proved to be effective in data augmentation. For DeBERTa, both GPT3-MIX \cite{yoo-etal-2021-gpt3mix-leveraging} and the proposed method employing soft labels performed significantly better than other methods. Although GPT3-MIX and the proposed method performed better on most datasets, the application of GPT3-MIX was limited for long sentences, such as in IMDb, due to the maximum token limit of the GPT3 API. The difference between the method proposed in this paper and GPT3-MIX is whether soft labels are assigned to randomly generated sentences or intentionally set soft labels. Additionally, if sentences are augmented through an LLM, the model relies heavily on the knowledge of the LLM. In other words, the unique characteristics of the dataset may become less prominent. However, as the proposed method focuses more on the given dataset, it is more effective in terms of this perspective because we can create soft labels without losing the characteristics of the dataset.

The proposed method exhibited a relatively small standard deviation of performance compared to other techniques, indicating that the proposed approach is statistically stable against changes in a random seed, unlike other augmentation methods. The construction of the low-resource dataset may vary with random seeds when randomly extracting data, leading to statistical instability. Nevertheless, the proposed method consistently increased performance through soft labels and mid-K sampling.

\subsection{Robustness}

\begin{table}[t]
\centering
\resizebox{\columnwidth}{!}{%
\begin{tabular}{c|ll}
\Xhline{3\arrayrulewidth}
\textbf{TextFooler} & \multicolumn{1}{c}{SUBJ}                                                                             & \multicolumn{1}{c}{IMDB}                                                                            \\ \Xhline{2\arrayrulewidth}
EDA                 & \begin{tabular}[c]{@{}l@{}}AUA: 22.0\%\\ ASR: 76.09\%\end{tabular} & \begin{tabular}[c]{@{}l@{}}AUA: 0.0\%\\ ASR: 100.0\%\end{tabular} \\ \hline
GPT3-MIX            & \begin{tabular}[c]{@{}l@{}}AUA: 8.0\%\\ ASR: 91.11\%\end{tabular}  & \begin{tabular}[c]{@{}l@{}}AUA: 2.0\%\\ ASR: 97.14\%\end{tabular} \\ \hline
Ours                & \begin{tabular}[c]{@{}l@{}}AUA: \textbf{36.0\%}\\ ASR: \textbf{60.0\%}\end{tabular}  & \begin{tabular}[c]{@{}l@{}}AUA: \textbf{6.0\%}\\ ASR: \textbf{92.68\%}\end{tabular} \\ \Xhline{3\arrayrulewidth}
\textbf{PWWS}       & \multicolumn{1}{c}{SUBJ}                                                                             & \multicolumn{1}{c}{IMDB}                                                                            \\ \Xhline{2\arrayrulewidth}
EDA                 & \begin{tabular}[c]{@{}l@{}}AUA: 30.0\%\\ ASR: 67.39\%\end{tabular} & \begin{tabular}[c]{@{}l@{}}AUA: 0.0\%\\ ASR: 100.0\%\end{tabular} \\ \hline
GPT3-MIX            & \begin{tabular}[c]{@{}l@{}}AUA: 22.0\%\\ ASR: 75.56\%\end{tabular} & \begin{tabular}[c]{@{}l@{}}AUA: 0.0\%\\ ASR: 100.0\%\end{tabular} \\ \hline
Ours                & \begin{tabular}[c]{@{}l@{}}AUA: \textbf{42.0\%}\\ ASR: \textbf{53.33}\%\end{tabular} & \begin{tabular}[c]{@{}l@{}}AUA:\ \textbf{4.0\%}\\ ASR: \textbf{95.12\%}\end{tabular} \\ \Xhline{3\arrayrulewidth}
\end{tabular}
}
\caption{Robustness against TextFooler and probability-weighted word saliency (PWWS) attacks on the baseline and proposed methods. AUA denotes accuracy under attack and ASR denotes attack success rate.}
\label{tab:robustness}
\end{table}

We experimented to evaluate the robustness against adversarial attacks, which induce changes in the predicted value of the model by altering several words. We employed the TextFooler \cite{jin2020bert} and probability-weighted word saliency \cite{ren-etal-2019-generating} strategies provided by the TextAttack \cite{morris-etal-2020-textattack} library. Following the previous approach \cite{si-etal-2021-better}, we evaluated the approach by selecting 10\% of the testing set from the SUBJ and IMDb datasets. We report accuracy under attack (AUA) and attack success rate (ASR) as metrics to assess the results. Table~\ref{tab:robustness} presents the results.

The experimental results demonstrate that the model trained using the proposed method exhibits higher robustness against adversarial attacks compared to other methods. The proposed method reduces overconfidence by generating data with a soft label that is close to the decision boundary for training \cite{muller2019does, thulasidasan2019Mixup}. Alleviating the overconfidence in this manner enhances the robustness of the model against adversarial attacks \cite{grabinski2022robust}. These findings are consistent with a previous study \cite{zhu-etal-2022-improving} demonstrating robustness improvement using data close to the decision boundary.

\subsection{Ablation Study}


\subsubsection{Effectiveness of Soft Labels}

We performed an ablation study to investigate the effectiveness of soft labels. This study compares the performance of forming augmented data pairs by assigning hard labels from the original sentences instead of using soft labels. Table~\ref{tab:hard_label} lists the results, confirming that the proposed method based on soft labels outperformed the approach based on assigning hard labels from the original data. These results suggest that soft labels play a crucial role in decision-boundary recognition augmentation methods. Additionally, using hard labels exhibited a higher standard deviation of performance than soft labels, indicating that this is more unstable than the soft label approach.

\subsubsection{Effectiveness of Curriculum Data Augmentation}

\begin{table}[t]
\centering
\resizebox{\columnwidth}{!}{%
\begin{tabular}{c|cc}
\Xhline{3\arrayrulewidth}
      & Baseline (Soft-Label) & Hard-Label            \\ \Xhline{2\arrayrulewidth}
SST2  & \textbf{88.9 (0.6)}   & 87.2 (1.2)            \\ \hline
SUBJ  & \textbf{86.7 (1.1)}   & 86.4 (5.0)            \\ \hline
TREC6 & \textbf{68.2 (4.1)}   & 51.3 (5.6)            \\ \Xhline{3\arrayrulewidth}
\end{tabular}
}
\caption{An ablation study on comparison between the soft-label and hard-label. We denote performance (\%) in mean (std) format.}
\label{tab:hard_label}
\end{table}
\setlength{\floatsep}{10pt} 

Recently, studies have explored the combination of textual data augmentation methods with curriculum learning \cite{bengio2009curriculum}, suggesting the concept of curriculum data augmentation \cite{wei-etal-2021-shot, lu-lam-2023-pcc}. The augmentation method proposed in this paper can also be extended to curriculum data augmentation by adjusting the number of gradient modifications. In this study, we employed a curriculum data augmentation approach, where data close to the decision boundary were gradually generated by involving data obtained through moving lambda $(\lambda)$ one more time every two epochs. The performance compared to the baseline is presented in Table~\ref{tab:cur}. For SST2, SUBJ, and CoLA, which consist of relatively short sentences, the curriculum augmentation showed a slight improvement at about 0.1 to 0.3. Additionally, it was not effective for dataset with long sentences such as IMDb. This result suggests that even though performance can be further improved with curriculum learning, its effectiveness is limited to the dataset with short sentences. However, since there was an improvement in performance in short sentences, it is expected that the improved approach will lead to performance enhancements in longer sentences as well. We leave this to future work.

\subsubsection{Effectiveness of Different Decoding Strategies}

\begin{table}[t]
\centering
\resizebox{\columnwidth}{!}{%
\begin{tabular}{c|cc}
\Xhline{3\arrayrulewidth}
     & \begin{tabular}[c]{@{}c@{}}Baseline (w/o Curr. Aug.)\end{tabular} & w/ Curr. Aug. \\ \Xhline{2\arrayrulewidth}
SST2 & \textbf{88.9 (0.6)}                                                              & \textbf{88.9 (1.7)}                 \\ \hline
SUBJ & 86.7 (1.1)                                                                       & \textbf{87.0 (4.8)}        \\ \hline
CoLA & 67.5 (2.0)                                                                       & \textbf{67.8 (1.1)}        \\ \hline
IMDB & \textbf{84.6 (5.7)}                                                              & 80.6 (2.8)                 \\ \Xhline{3\arrayrulewidth}
\end{tabular}
}
\caption{An ablation study on curriculum augmentation. We denote performance (\%) in mean (std) format.}
\label{tab:cur}
\end{table}
\setlength{\textfloatsep}{10pt} 

\begin{table}[t]
\centering
\resizebox{\columnwidth}{!}{%
\begin{tabular}{c|cccc}
\Xhline{3\arrayrulewidth}
     & Greedy      & Beam Search & Top-K      & Mid-K               \\ \Xhline{2\arrayrulewidth}
SST2 & 87.6 (2.5)  & 86.5 (2.7)  & 88.6 (0.7) & \textbf{88.9 (0.6)} \\ \hline
SUBJ & 73.6 (12.3) & 81.3 (6.0)  & 86.3 (1.0) & \textbf{86.7 (1.1)} \\ \hline
CoLA & 63.9 (4.3)  & 67.4 (1.8)  & 66.4 (0.8) & \textbf{67.5 (2.0)} \\ \Xhline{3\arrayrulewidth}
\end{tabular}
}
\caption{An ablation study on decoding strategy. We denote the performance (\%) of our proposed Mid-K sampling and other decoding methods in mean (std) format.}
\label{tab:decoding}
\end{table}

\begin{table*}[t]
\resizebox{\textwidth}{!}{%
\begin{tabular}{c|c}
\Xhline{3\arrayrulewidth}
Original                                                                                                                                                                                                                                                                                                                                                                                                               & Augmented                                                                                                                                                                                                                                                                                                                                                                                                \\ \Xhline{2\arrayrulewidth}
\multicolumn{1}{l|}{a dark, quirky road movie that constantly \textbf{defies expectation.}}                                                                                                                                                                                                                                                                                                                                     & \multicolumn{1}{l}{a has all but \textbf{no sense of story but fascinating humor.}}                                                                                                                                                                                                                                                                                                                               \\
(Positive: 75\% / Negative 25\%)                                                                                                                                                                                                                                                                                                                                                                                       & (Positive: 100\% / Negative 0\%)                                                                                                                                                                                                                                                                                                                                                                         \\ \hline
\multicolumn{1}{l|}{\begin{tabular}[c]{@{}l@{}}This film contains more action before the opening credits than\\ are in entire Hollywood films of this sort. This film is produced\\  by Tsui Hark and stars Jet Li. This team has brought you many\\  worthy \textbf{Hong Kong} cinema productions, including the Once\\  Upon A Time in China series. \textbf{The action was fast and furious}\\  with amazing wire work.\end{tabular}} & \multicolumn{1}{l}{\begin{tabular}[c]{@{}l@{}}This film begins more action before the opening credits than\\ are in entire Hollywood films of this sort. This film is based\\ byeki Hol and stars Michael Ryan. This team has made you\\ many wonderful \textbf{Hollywood} cinema productions, including Most\\ Lear Aot in Friday The Wellle car series. \textbf{The action, the story}\\ \textbf{actually was dry.}\end{tabular}} \\
(Positive: 100\% / Negative 0\%)                                                                                                                                                                                                                                                                                                                                                                                       & (Positive: 74\% / Negative 26\%)                                                                                                                                                                                                                                                                                                                                                                         \\ \hline
\multicolumn{1}{l|}{\begin{tabular}[c]{@{}l@{}}These thoughts are \textbf{hugely entertaining} and women will\\ also enjoy this movie I’m sure! All cast members perform well,\\ and this film could have been a tremendous hit all over the world\\ if it was made in \textbf{England or the US.} But for those of you\\ who are fortunate enough to understand Swedish, \textbf{you are}\\ \textbf{in for a treat!}\end{tabular}}                        & \multicolumn{1}{l}{\begin{tabular}[c]{@{}l@{}}Some brains are \textbf{terrific} and women will also enjoy this movie\\ I’m sure! First cast members plays well, and this film could have\\ been a particularly look all over the world if it was made\\ in \textbf{Hollywood or the era.} But for those of you who are complicit\\ enough to Aotta, \textbf{you are in for hatred!}\end{tabular}}                                   \\
(Positive: 100\% / Negative 0\%)                                                                                                                                                                                                                                                                                                                                                                                       & (Positive: 75\% / Negative 25\%)                                                                                                                                                                                                                                                                                                                                                                         \\ \Xhline{3\arrayrulewidth}
\end{tabular}
}
\caption{Augmented samples with its soft-label through our proposed method. The left column of each row denotes the original data, and the right column represents the augmented data through our method. Important differences between original data and augmented data are boldfaced.}
\label{tab:case}
\end{table*}

We conducted an ablation study to validate the effectiveness of mid-K sampling compared to other conventional decoding strategies. Table~\ref{tab:decoding} presents the results of applying greedy decoding, the beam search, and top-K sampling to the proposed method, compared to the baseline method that uses mid-K sampling. Specifically, applying greedy decoding or the beam search resulted in lower performance compared to mid-K sampling. Greedy decoding and beam search are optimization methods that aim to determine an optimal solution, making them effective in such tasks as translation. Top-K sampling introduces relatively more diversity through sampling compared to the greedy method; however, the produced sentences still frequently resemble the original because this method prefers words in the original sentences with the highest probability. In contrast, the mid-K sampling proposed in this paper achieves the highest performance because it provides diversity while effectively preserving semantic coherence.

\subsection{Case Study}

Table~\ref{tab:case} presents augmented sentences through the proposed method. Three main characteristics of the proposed method are observed by comparing the original and augmented data. 

First, the method proposed in this paper introduces different proper nouns compared to the original data. For instance, it changed “Hong Kong cinema productions” to “Hollywood cinema productions.” This transformation allows the model to learn from augmented data that are distinct from the original, serving as new and diverse training examples. This outcome distinguishes the proposed approach from such methods as word-level modification or back-translation, which only replace a few words in each sentence, resulting in augmented sentences that are similar to the original.

Second, the proposed method generates data close to the decision boundary while maintaining the core attribute. For example, through the example of “no sense of story but fascinating humor,” the proposed method produces sentences close to the decision boundary by adding the aspect of “no sense of story” while preserving the core attribute of “fascinating humor.” This characteristic allows us to leverage data with soft labels that are close to the decision boundary, preventing overconfidence in the model and improving performance and adversarial robustness.

Finally, the suggested approach allows for phrase-level modification of the expression and the generation of various expressions. For instance, applying a rule-based synonym replacement to “hugely entertaining” may result in such expressions as “very entertaining” or “hugely humorous.” However, the proposed approach can produce expressions like “terrific.” This characteristic facilitates the training of the model with a variety of expressions by allowing for a broader range of expressions beyond simple word-level modifications. Furthermore, the word modifications take place at a level that is uncommon but may still be inferred from their meaning, such as changing “thoughts” to “brains.” By employing such exceptional modifications, as opposed to relying on predefined dictionaries, the proposed model can improve its robustness against an adversarial attack that confuses the model with unexpected expressions or expressions the model might not have learned well.

\section{Conclusion}

This paper introduces a novel text augmentation method aimed at enhancing model robustness by shifting features closer to the decision boundary and increasing data diversity through mid-K sampling. Experimental results demonstrate the efficacy of decision-boundary-aware soft labels and mid-K sampling in augmenting data diversity and robustness. While the proposed approach shows promise, there are potential limitations, particularly related to covariate shift between augmented and real-world data distributions. Future research should focus on quantifying and mitigating this shift to ensure the augmented data's representativeness.

\section*{Limitations}

While our paper introduces a novel augmentation technique with various benefits, we acknowledge several limitations:

\begin{itemize}
    \item \textbf{Concerns on Covariate Shift}: The proposed augmentation method assumes alignment between augmented and real-world data distributions, potentially introducing covariate shift. Future research would address strategies to quantify and mitigate this shift, especially in low-resource settings.

    \item \textbf{Linguistic Correctness}: Generated sentences may lack linguistic correctness, albeit previous studies suggest performance enhancement despite linguistic inaccuracies. Our mid-K sampling approach primarily modifies proper nouns, preserving linguistic structure and semantics while introducing diversity. Additionally, adjusting hyperparameters and incorporating soft-label adaptation aids to mitigate linguistic damage.

    \item \textbf{Curriculum Data Augmentations}: While curriculum augmentation was not significantly effective in this study, its potential benefits, especially in specific datasets, suggest further exploration. Future work will focus on refining curriculum augmentation strategies to improve effectiveness and robustness.
\end{itemize}

\section*{Ethics Statement}

The proposed method relies on pretrained language models to generate sentences. This dependence on specific models leads to potential bias in the augmented sentences. However, as the proposed method aims to move the given representation close to the decision boundary in the feature space, resulting in weakening strong expressions, it may help neutralize biased expressions in the original data.

\section*{Acknowledgements}
This research was supported by Basic Science Research Program through the National Research Foundation of Korea(NRF) funded by the Ministry of Education(NRF-2022R1C1C1008534), and Institute for Information \& communications Technology Planning \& Evaluation (IITP) through the Korea government (MSIT) under Grant No. 2021-0-01341 (Artificial Intelligence Graduate School Program, Chung-Ang University).

\section*{Bibliographical References}
\bibliographystyle{lrec_natbib}
\bibliography{lrec-coling2024-example}

\clearpage

\appendix

\noindent
\begin{minipage}[!htbp]{\textwidth}
\centering
    \begin{minipage}[h]{\textwidth}
        \centering
        \resizebox{\textwidth}{!}{
        \begin{tabular}{c|c}
        \Xhline{3\arrayrulewidth}
        Original Sentence & \begin{tabular}[c]{@{}c@{}}This movie followed movies within a movie, much like Scream 3 and Urban Legend 2.\\ This was pure crap! The whole movie within a movie crap.\end{tabular} \\ \hline
        Fine-tuned BERT   & \begin{tabular}[c]{@{}c@{}}This movie followed movies within a movie, much like Scream 3 and Urban Legend 2.\\ This was \textbf{pure} crap! The whole movie within a movie crap.\end{tabular} \\ \hline
        Ours              & \begin{tabular}[c]{@{}c@{}}This movie followed movies within a movie, much like Scream 3 and Urban Legend 2.\\ This was pure \textbf{crap}! The whole movie within a movie \textbf{crap}.\end{tabular} \\ \Xhline{3\arrayrulewidth}
        \end{tabular}%
        }
        \captionof{table}{Examples for case study in robustness. The bolded words represent important terms selected via Lime~\cite{ribeiro2016should} for the given model.}
        \label{tab:my-table7}

    \end{minipage}
\end{minipage}


\section{Case Study in Robustness}
We investigated which words in the sentence contained the adversarial attack in order to evaluate the model's robustness using examples. The sentence chosen from IMDB for investigation was, "This movie followed movies within a movie, much like Scream 3 and Urban Legend 2." This was really terrible! The entire film within a film is crap.". This sentence has negative sentiment owing to expressions such as "crap". Table~\ref{tab:my-table7} presents augmented sentences through the proposed method.

However, when we applied adversarial attack to the fine-tuned BERT model trained on original data using this sentence, TextFooler attack algorithm changed the word "pure" instead of the word "crap", which actually contributes the negative sentiment of the sentence. For further investigation, we examined the word importance of the model on this sentences, and discovered that "pure" had a high level of importance according to the fine-tuned BERT model. Nonetheless, in the BERT model trained on original data and augmented data with our method, the word "crap" received a high level of word importance. Moreover, the model successfully defended the adversarial attack on this example sentence, different from the model trained only on original data. We concluded that this correction of word importance enhances the robustness of the model.


\end{document}